# Generalized Non-orthogonal Joint Diagonalization with LU Decomposition and Successive Rotations

Xiao-Feng Gong, *Member, IEEE*, Xiu-Lin Wang, and Qiu-Hua Lin, *Member, IEEE*

***Abstract*—Non-orthogonal joint diagonalization (NJD) free of prewhitening has been widely studied in the context of blind source separation (BSS) and array signal processing, etc. However, NJD is used to retrieve the jointly diagonalizable structure for a single set of target matrices which are mostly formulized with a single dataset, and thus is insufficient to handle multiple datasets with inter-set dependences, a scenario often encountered in joint BSS (J-BSS) applications. As such, we present a generalized NJD (GNJD) algorithm to simultaneously perform asymmetric NJD upon multiple sets of target matrices with mutually linked loading matrices, by using LU decomposition and successive rotations, to enable J-BSS over multiple datasets with indication / exploitation of their mutual dependences. Experiments with synthetic and real-world datasets are provided to illustrate the performance of the proposed algorithm.**

## I. Introduction

Joint diagonalization (JD) is an important instrument in solving blind source separation (BSS) problems. For example, consider an instantaneous linear mixture $\boldsymbol{x}(t) = \boldsymbol{As}(t) \in \mathbb{C}^N$, where $\boldsymbol{s}(t) \in \mathbb{C}^N$ and $\boldsymbol{A} \in \mathbb{C}^{N\times N}$ denote the source and mixing matrix, respectively. The sources are assumed mutually independent, or uncorrelated but with some temporal structures (e.g. non-stationarity, non-whiteness). We can then calculate the 4th-order cumulant [1] or 2nd-order covariance matrices (at distinct time instants or time shifts) [2-4] $\boldsymbol{C}_1,\ldots,\boldsymbol{C}_K$ under the above assumptions, that share the following jointly diagonalizable structure:

$$\boldsymbol{C}_k = \boldsymbol{A}\boldsymbol{D}_k\boldsymbol{A}^H \quad (1)$$

where $\boldsymbol{D}_k$ is diagonal, $k = 1,\ldots,K$, and superscript '$H$' denotes conjugated transpose. JD then seeks an estimate of $\boldsymbol{A}$ by fitting the above common JD structure.

Numerous algorithms for computing JD were proposed in the open literature. While the early works are mostly focused on orthogonal JD (OJD) which requires $\boldsymbol{A}$ to be unitary and thus applied in BSS of pre-whitened mixtures [1, 2], the recent efforts, on the other hand, turned to non-orthogonal JD (NJD) to facilitate BSS free of pre-whitening for both real-valued and complex-valued mixtures [3-14]. Criteria including weighted least squares [4-8], minimization of off-norm [9-13], and information theoretic criterion [14] were successfully adopted, that are specifically accomplished via several optimization strategies such as Gauss-Newton [8] and successive rotations [10-14].

Although JD has found growing interests in both theory and application, the majority of its contributions are with regards to single-set data analysis such as BSS of a single set of linear instantaneous mixtures. More precisely, the target matrices for JD are mostly formulized by computing the intra-set statistics (e.g. auto-covariance, 4th-order cumulant) of a single dataset, and are thus with symmetric or Hermitian structure. In this background, most of the JD works were historically devised for symmetric or Hermitian target matrices whose row and column spaces are identical [1-14]. However, when multiple datasets with inter-set dependences are available, for example, when working on multi-subject/multi-modal biomedical data fusion problems [15, 20-23], or BSS of transformed signals in multiple frequency bins [16], the use of NJD fails to sufficiently utilize these inter-set dependences, and this will explicitly result in problems such as permutation misalignment or loss of accuracy [24, 25]. As such, generalized JD (GJD), that incorporates both intra-set and inter-set statistics to enable joint BSS (J-BSS) of multi-set data, has become an issue of great interests. Indeed, although J-BSS has already been addressed in other aspects of BSS, for example, in independent vector analysis (IVA) [15-17], canonical correlation analysis (CCA) [18, 19, 22], and multi-set CCA (MCCA) [20-22], efforts with GJD towards J-BSS are still limited [24 - 29]. The main difficulty lies in the fact that calculating inter-set statistics brings asymmetric target matrices for GJD, yet the derivations for most of the existing JD works were devised for symmetric or Hermitian problems.

More precisely, asymmetric OJD or joint SVD (J-SVD) was studied in [25], which could be used for J-BSS of 2 pre-whitened datasets. The work in [24, 26] considered generalized OJD (GOJD) problem, to facilitate J-BSS of 3 or more pre-whitened datasets. The authors of [26] also considered for the first time non-orthogonal joint BSS (NOJoB) for 3 or more datasets within the generalized NJD (GNJD) context. This algorithm was originally devised for real-valued problems, and could be extended to the complex case with some tiny modifications. We also considered in [27] the joint solution of multiple asymmetric NJD problems as a preliminary work of the presented one.

This work was supported in part by Doctoral Fund of Ministry of Education of China under grant 20110041120019, National Natural Science Foundation of China under grants 61072098, 61105008, 61331019, 61379012, and Scientific Research Fund of Liaoning Provincial Education Department under grant L2014016.

Xiao-Feng Gong, Xiu-Lin Wang and Qiu-Hua Lin are with the School of Information and Communication Engineering, Dalian University of Technology, Dalian, China, 116024 (e-mails: xfgong@dlut.edu.cn; wxl1482@mail.dlut.edu.cn; qhlin@dlut.edu.cn)

However, the joint NJD (JNJD) algorithm therein makes use of dependences between adjacent datasets only, and thus still has problems when handling large number of datasets. In addition, several GJD algorithms have been successfully applied to constrained or coupled tensor factorization problems [28, 29]. More precisely, the canonical polyadic decomposition (CPD) with constraints of constant modulus was converted into a simple GJD formulation in [28], such that 2 sets of target matrices (one set is Hermitian and the other is symmetric) with a common loading matrix are handled simultaneously. In [29], we considered coupled CPD of 2 tensors with a shared loading matrix and demonstrated how it could be solved with the JNJD algorithm developed in [27].

In this study, we propose another GNJD algorithm for the joint analysis of multiple datasets. More specifically, the GNJD problem is set up with multiple asymmetric NJD problems, of which every two distinct ones are linked by a shared loading matrix. By exploiting these mutual links across datasets, the multiple asymmetric NJD problems are solved simultaneously using LU decomposition and successive elementary rotations. The proposed GNJD algorithm relieves the orthogonality constraints for J-SVD and GOJD, and could be used to handle 3 or more datasets. In addition, when compared with NOJoB, we note that similar problem is considered but with distinct optimization strategies. In particular, the proposed GNJD algorithm yields better convergence behavior when handling large number of datasets, and improved performance in high noise levels, as will be shown later.

The rest of the paper is organized as follows. In Section II, we give the formulization of the GNJD problem and some examples on how practical problems could be linked to GNJD. In Section III, we present the proposed algorithm as well as theoretical analysis including computational complexity and convergence, and some implementation remarks. Experiment results are given in Section IV to illustrate the performance of the proposed algorithm. Finally, Section V concludes this paper. The source programs for the proposed algorithm are available at [39].

## II. Problem Formulation

In this section, we present the data model for multi-set processing and further give examples on how the multi-set data model could be formulized into GNJD problems. In addition, comparisons with existing JD and GJD formulizations are also given to provide insights into the GNJD model.

### A. Multi-set data model

Recently, multi-set data processing has attracted much attention in the literature [15-27]. The key idea to these works is to incorporate both intra-set independence (as is assumed in classical BSS problems) and inter-set dependence at the source level, to achieve J-BSS or data fusion for multiple datasets. The following multi-set instantaneous mixing model is assumed:

$$\boldsymbol{x}_r(t) = \boldsymbol{A}_r \boldsymbol{s}^{(r)}(t), \qquad r = 1,2,\ldots,R \tag{2}$$

where $\boldsymbol{x}_r(t)$, $\boldsymbol{s}^{(r)}(t) \in \mathbb{C}^N$, $\boldsymbol{A}_r \in \mathbb{C}^{N\times N}$ denote the observation, source, and mixing matrix in the *rth* dataset, respectively, $r = 1,2,\ldots,R$. By defining new source vectors:

$$\boldsymbol{s}_n(t) \triangleq [s_n^{(1)}(t),\ldots,s_n^{(R)}(t)]^T \in \mathbb{C}^R, \qquad n = 1,\ldots,N \tag{3}$$

we note that $\boldsymbol{s}_n(t)$ and $\boldsymbol{s}_m(t)$ are independent for any $1 \le m \ne n \le N$ (intra-set independence), and that components of $\boldsymbol{s}_n(t)$ are mutually dependent (inter-set dependence).

The above multi-set data model has been largely considered in practical problems. For example, in BSS of convolutive mixtures, (2) models the linear mixing procedure at the *rth* frequency bin, and the inter-set dependence mentioned above formulizes the well-known cross frequency dependences that are extensively used in frequency domain BSS of convolutive mixtures [16].

The above model is also widely used in multi-set data fusion with emphasis on finding their similarities or connections, and has found applications in biomedical engineering [15, 20-23]. For example, the multiple datasets might refer to data of different modalities (e.g. fMRI, EEG, sMRI) collected from a single subject under equal conditions, and joint analysis methods that simultaneously decompose these datasets with indications of their relations are of high interests [21, 22]. Moreover, the multiple datasets could as well refer to those collected from multiple subjects under identical modality and conditions. Typical examples include multi-subject fMRI data and hyper-scanning EEG data [20, 23]. In addition, the above multi-set data model was also used in array processing applications to formulate signals collected from distinct sensors (e.g. electrocardiogram data collected with multi-electrodes [24]), or array statistics of distinct orders or forms (e.g. covariance and pseudo-covariance matrices [29]).

### B. GNJD formulation

The GNJD formulation contains $(R+1)R/2$ sets of NJD problems (mostly asymmetric) of the following form:

$$\boldsymbol{C}_{r_1,r_2,k} \triangleq \boldsymbol{A}_{r_1} \cdot \boldsymbol{D}_{r_1,r_2,k} \cdot \boldsymbol{A}_{r_2}^H \tag{4}$$

where $\boldsymbol{A}_{r_1}, \boldsymbol{A}_{r_2} \in \mathbb{C}^{N\times N}$ denote loading matrices for the $r_1 th$ and $r_2 th$ datasets, $\boldsymbol{C}_{r_1,r_2,k}, \boldsymbol{D}_{r_1,r_2,k} \in \mathbb{C}^{N\times N}$, denote the target matrices and the unloaded diagonal matrices, respectively, $k = 1,\ldots,K$, $1 < r_1 \le r_2 < R$. Integers $R$, $K$ and $N$ denote the number of loading matrices, number of target matrices in each NJD dataset, and dimensionality of target matrices, respectively. We note that the $(r_1, r_2)th$ set of target matrices share with the $(r_1, l)th$ and $(l', r_2)th$ sets the loading matrices $\boldsymbol{A}_{r_1}$ and $\boldsymbol{A}_{r_2}$, respectively, $l = r_1,\ldots,R$, $l' = 1,\ldots,r_2$. Therefore, all these $(R+1)R/2$ NJD problems are mutually connected with one another. GNJD then aims at estimating all the $R$ loading matrices $\boldsymbol{A}_1, \boldsymbol{A}_2, \ldots, \boldsymbol{A}_R$ such that the jointly diagonalizable structures for all the $(R+1)R/2$ NJD problems in (4) are fitted simultaneously.

It is important to note that the above GNJD formulation could be derived from the multi-set data model provided in subsection II. A. For example, with source non-stationarity present we could calculate the following target matrices:

$$\boldsymbol{C}_{r_1,r_2,k} \triangleq \mathrm{E}\{\boldsymbol{x}_{r_1}(k)[\boldsymbol{x}_{r_2}(k)]^H\} = \boldsymbol{A}_{r_1}\, \mathrm{E}\{\boldsymbol{s}^{(r_1)}(k)[\boldsymbol{s}^{(r_2)}(k)]^H\} \boldsymbol{A}_{r_2}^H \tag{5}$$

where $r_1, r_2 = 1,2,\ldots,R$. Moreover, we note here that $\mathrm{E}\{\boldsymbol{s}^{(r_1)}(k)\cdot[\boldsymbol{s}^{(r_2)}(k)]^H\}$ is diagonal under the basic assumptions of J-BSS of intra-set independence and inter-set dependence, and thus (5) is actually a GNJD formulation. In addition, the GNJD formulization could also be obtained by calculating the cross 4th-order cumulants upon the multi-set data model, as is done in [24].

### C. Comparison with existing formulations for BSS and J-BSS

The considered GNJD model is distinct from the NJD models as well as some other linear algebraic models that are well-established for BSS and J-BSS. More exactly, we note that the majority of NJD works considered symmetric or Hermitian target matrices [1-14], which were usually constructed with 2nd-order or 4th-order intra-set statistics (such as covariance) of single dataset observations[1]. Therefore, the NJD works are mostly devoted to BSS of a single dataset, and the reasonings therein might be invalid for asymmetric NJD problems (in the sense that the row and column spaces are distinct) involved in the GNJD formulation.

For the joint analysis of 2 datasets, some linear algebraic models have been established by calculating inter-set statistics across the 2 datasets. The classical CCA method is one such example, for which the cross-covariance matrix of 2 datasets is established as the sole target matrix [18, 19, and 22]. The work in [25] further extends the above one sample cross-covariance model in CCA into multiple samples to form an asymmetric JD (or J-SVD) model. It is important to note that the above CCA and asymmetric JD models could be considered as special cases of GNJD. More exactly, GNJD model in (4) degrades to the asymmetric JD model if we let $R = 2$, $r_1 = 1$ and $r_2 = 2$, which is further reduced to CCA if we set $K$ to 1.

Recently, some advanced models were developed for multiple datasets (more than 2). In particular, the MCCA model extends CCA so as to include one sample cross-covariance matrices among every 2 pre-whitened datasets [20-22]. The GOJD model considered in [24] further extends MCCA to the multi-sample case. In particular, GNJD model could be converted into GOJD if we require the loading matrices $\boldsymbol{A}_r$'s to be unitary, $r = 1, 2, \ldots, R$, and that $r_2$ be strictly larger than $r_1$, and GOJD could be further converted into MCCA if we set $K$ to 1. In addition, GNJD is also different from the JNJD formulation in [27], if we note that JNJD only considered a subset of GNJD model with $r_2 = r_1 + 1$.

As a result, GNJD is a more generalized model for multi-set data analysis than the existing CCA, asymmetric JD, MCCA, GOJD, and JNJD models. In addition, we note that the above models for multi-set analysis are distinct from ordinary NJD in that the latter fails to consider the inter-set dependences. In particular, advantages of using multi-set models over NJD have been partially addressed in [25, 27], with respects to permutation alignment and estimation accuracy.

To our best knowledge, the above GNJD model has only been considered once in the open literature [26], where a non-orthogonal joint BSS (NOJoB) algorithm was proposed based on power iterations. However, this algorithm will occasionally suffer from non-optimal converging patterns and performance loss in highly noisy environment, as will be shown in the experiment section. In the next section, we propose an algorithm to solve the GNJD problem with LU decomposition and successive rotations.

[1] Sometimes the target matrices may neither be symmetric nor Hermitian such as the complex time-lagged covariance matrices [2]. We note that the row and column spaces of these target matrices are identical, and thus the target matrices could be converted into Hermitian by adding with their conjugated transposes.

## III. Proposed Algorithm

In this section, we illustrate how the GNJD problem is solved with LU decomposition and successive elementary rotations. In addition, we present some analysis and implementation remarks to provide insights into the proposed algorithm.

### A. GNJD with LU decomposition and successive rotations

To solve the GNJD problem, we extend the well-established off-norm minimization criterion [9-13] to GNJD as follows:

$$\{\boldsymbol{B}_1,\ldots,\boldsymbol{B}_R\} = \underset{\boldsymbol{B}_1,\ldots,\boldsymbol{B}_R}{\arg\min} \sum_{k=1}^{K}\sum_{r_2=r_1}^{R}\sum_{r_1=1}^{R}\left\|\mathrm{off}(\boldsymbol{B}_{r_1}\cdot\boldsymbol{C}_{r_1,r_2,k}\cdot\boldsymbol{B}_{r_2}^H)\right\|_F^2 \quad (6)$$

where $\boldsymbol{B}_r \triangleq \boldsymbol{A}_r^{-1}$ denotes the $rth$ unloading matrix, $\mathrm{off}(\cdot)$ is the operator that sets all the diagonal elements of its entry to zero, and $\|\cdot\|_F$ is Frobenius norm.

We note that the problem in (6) involves lots of parameters to adjust during the optimization. Therefore, we adopt LU decomposition and successive rotation based scheme to ease this problem into loops over simple linear sub-optimizations [1, 10-14, 27]. More exactly, we consider the LU decomposition of $\boldsymbol{B}_r$, $r = 1, 2, \ldots, R$:

$$\boldsymbol{B}_r = \boldsymbol{L}_r\boldsymbol{U}_r \quad (7)$$

where $\boldsymbol{L}_r, \boldsymbol{U}_r \in \mathbb{C}^{N\times N}$ are the $rth$ lower-triangular and upper-triangular matrices, respectively, $r = 1, \ldots, R$, and then (6) could be converted into the following 2 alternating stages, which update either $\boldsymbol{L}_r$ or $\boldsymbol{U}_r$ one at a time:

$$\{\tilde{\boldsymbol{U}}_r \mid r = 1,\ldots,R\} = \underset{\tilde{\boldsymbol{U}}_1,\ldots,\tilde{\boldsymbol{U}}_R}{\arg\min} \sum_{k=1}^{K}\sum_{r_2=r_1}^{R}\sum_{r_1=1}^{R}\left\|\mathrm{off}(\tilde{\boldsymbol{U}}_{r_1}\boldsymbol{C}_{r_1,r_2,k}\tilde{\boldsymbol{U}}_{r_2}^H)\right\|_F^2 \quad (8)$$

$$\{\tilde{\boldsymbol{L}}_r \mid r = 1,\ldots,R\} = \underset{\tilde{\boldsymbol{L}}_1,\ldots,\tilde{\boldsymbol{L}}_R}{\arg\min} \sum_{k=1}^{K}\sum_{r_2=r_1}^{R}\sum_{r_1=1}^{R}\left\|\mathrm{off}(\tilde{\boldsymbol{L}}_{r_1}\boldsymbol{C}'_{r_1,r_2,k}\tilde{\boldsymbol{L}}_{r_2}^H)\right\|_F^2 \quad (9)$$

where $\boldsymbol{C}'_{r_1,r_2,k} = \tilde{\boldsymbol{U}}_{r_1}\boldsymbol{C}_{r_1,r_2,k}\tilde{\boldsymbol{U}}_{r_2}^H$, $\tilde{\boldsymbol{U}}_r, \tilde{\boldsymbol{L}}_r$ denote the updates of $\boldsymbol{U}_r, \boldsymbol{L}_r$. In the proposed algorithm, (8) and (9) are alternated until convergence is reached. Next we shall explain how these 2 sub-optimization problems could be solved using successive rotations. Before we start, we note that these 2 stages are similar as they both seek estimates for triangular matrices, and thus we shall only discuss the L-stage in (9).

More exactly, the updating of $\{\tilde{\boldsymbol{L}}_r \mid r = 1,\ldots,R\}$ is in terms of successive products of elementary rotation matrices, each associated with an index pair $(i, j)$, by repeatedly solving the following sub-optimization problems for all index pairs $(i, j)$:

$$\begin{cases}\{\boldsymbol{T}_{r,(i,j)} \mid r = 1,\ldots,R\} = \underset{\boldsymbol{T}_1,\ldots,\boldsymbol{T}_R}{\arg\min} \sum\limits_{k=1}^{K}\sum\limits_{r_2=r_1}^{R}\sum\limits_{r_1=1}^{R}\left\|\mathrm{off}(\boldsymbol{T}_{r_1,(i,j)}\boldsymbol{C}_{r_1,r_2,k,old}\boldsymbol{T}_{r_2,(i,j)}^H)\right\|_F^2 \\ \boldsymbol{C}_{r_1,r_2,k,new} = \boldsymbol{T}_{r_1,(i,j)}\boldsymbol{C}_{r_1,r_2,k,old}\boldsymbol{T}_{r_2,(i,j)}^H, \qquad \boldsymbol{L}_{r,new} = \boldsymbol{T}_{r,(i,j)}\boldsymbol{L}_{r,old}\end{cases} \quad (10)$$

where $\boldsymbol{T}_{r,(i,j)}$ is the elementary rotation matrix associated with index pair $(i,j)$. $\boldsymbol{C}_{r_1,r_2,k,new}$, $\boldsymbol{L}_{r,new}$ are the most recent updates for $\boldsymbol{C}_{r_1,r_2,k}$ and $\boldsymbol{L}_r$, and $\boldsymbol{C}_{r_1,r_2,k,old}$, $\boldsymbol{L}_{r,old}$ are updates in the previous iteration. We note that $\boldsymbol{T}_{r,(i,j)}$ is an elementary lower triangular matrix defined as:

$$
\boldsymbol{T}_{r,(i,j)} \triangleq \begin{bmatrix} 1 & & & & \\ & \ddots & & & \\ & & 1 & & \\ & \alpha_{r,(i,j)} & & \ddots & \\ & & & & 1 \end{bmatrix} \tag{11}
$$

where the only non-zero off-diagonal element $\alpha_{r,(i,j)}$ is at the *ith* row and *jth* column. As such, we only need to estimate 2 sets of parameters $\{\alpha_{r_1,(i,j)} \mid r_1 = 1,\dots,R\}$ and $\{\alpha_{r_2,(i,j)} \mid r_2 = r_1,\dots,R\}$ by minimizing $\zeta_{(i,j),new} = \sum_{k=1}^{K}\sum_{r_2=r_1}^{R}\sum_{r_1=1}^{R} \| \mathrm{off}(\boldsymbol{T}_{r_1,(i,j)}\boldsymbol{C}_{r_1,r_2,k,old}\boldsymbol{T}^H_{r_2,(i,j)})\|_F^2$ in each iteration, formulized as:

$$
\begin{aligned}
\zeta_{(i,j),new} &= \sum_{r_2=r_1}^{R}\sum_{r_1=1}^{R}\sum_{k=1}^{K}\sum_{p=1,p\neq i}^{N} \left(|\boldsymbol{C}_{r_1,r_2,k,new}(i,p)|^2 + |\boldsymbol{C}_{r_1,r_2,k,new}(p,i)|^2\right) \\
&= \sum_{r=1}^{R}\sum_{k=1}^{K}\sum_{p=1,p\neq i}^{N} \left(\sum_{r_2=r}^{R} |\boldsymbol{C}_{r,r_2,k,new}(i,p)|^2 + \sum_{r_1=1}^{r} |\boldsymbol{C}_{r_1,r,k,new}(p,i)|^2\right)
\end{aligned} \tag{12}
$$

by denoting $\boldsymbol{C}_{r_1,r_2,k,new} = \boldsymbol{T}_{r_1,(i,j)}\boldsymbol{C}_{r_1,r_2,k,old}\boldsymbol{T}^H_{r_2,(i,j)}$ and noting that multiplying $\boldsymbol{C}_{r_1,r_2,k,old}$ from the left and right with $\boldsymbol{T}_{r_1,(i,j)}$ and $\boldsymbol{T}^H_{r_2,(i,j)}$ only impact its *ith* row and column, respectively.

Moreover, with some simple operations we could express $|\boldsymbol{C}_{r_1,r_2,k,new}(i,p)|^2$ and $|\boldsymbol{C}_{r_1,r_2,k,new}(p,i)|^2$ explicitly with $\alpha_{r_1,(i,j)}$ and $\alpha_{r_2,(i,j)}$ as follows:

$$
\begin{aligned}
|\boldsymbol{C}_{r_1,r_2,k,new}(i,p)|^2 = [&|\boldsymbol{C}_{r_1,r_2,k,old}(j,p)|^2\, \alpha_{r_1,(i,j)}\alpha^*_{r_1,(i,j)} \\
&+\boldsymbol{C}_{r_1,r_2,k,old}(i,p)\boldsymbol{C}^*_{r_1,r_2,k,old}(j,p)\alpha^*_{r_1,(i,j)} \\
&+\boldsymbol{C}_{r_1,r_2,k,old}(j,p)\boldsymbol{C}^*_{r_1,r_2,k,old}(i,p)\alpha_{r_1,(i,j)} + |\boldsymbol{C}_{r_1,r_2,k,old}(i,p)|^2]
\end{aligned} \tag{13}
$$

$$
\begin{aligned}
|\boldsymbol{C}_{r_1,r_2,k,new}(p,i)|^2 = [&|\boldsymbol{C}_{r_1,r_2,k,old}(p,j)|^2\, \alpha_{r_2,(i,j)}\alpha^*_{r_2,(i,j)} \\
&+\boldsymbol{C}_{r_1,r_2,k,old}(p,j)\boldsymbol{C}^*_{r_1,r_2,k,old}(p,i)\alpha^*_{r_2,(i,j)} \\
&+\boldsymbol{C}_{r_1,r_2,k,old}(p,i)\boldsymbol{C}^*_{r_1,r_2,k,old}(p,j)\alpha_{r_2,(i,j)} + |\boldsymbol{C}_{r_1,r_2,k,old}(p,i)|^2]
\end{aligned} \tag{14}
$$

Substituting (13) and (14) into (12) yields:

$$
\begin{aligned}
\zeta_{(i,j),new} = \sum_{r=1}^{R}\sum_{k=1}^{K}\sum_{p=1,p\neq i}^{N} [&a_{j,p,r}|\alpha_{r,(i,j)}|^2 + b_{i,j,p,r}\alpha^*_{r,(i,j)} \\
&+ b^*_{i,j,p,r}\alpha_{r,(i,j)} + a_{i,p,r}]
\end{aligned} \tag{15}
$$

where $a_{j,p,r}$, $b_{i,j,p,r}$ are defined as follows:

$$
\begin{cases}
a_{j,p,r} \triangleq \sum_{r_2=r}^{R} |\boldsymbol{C}_{r,r_2,k,old}(j,p)|^2 + \sum_{r_1=1}^{r} |\boldsymbol{C}_{r_1,r,k,old}(p,j)|^2 \\
b_{i,j,p,r} \triangleq \sum_{r_2=r}^{R} \boldsymbol{C}_{r,r_2,k,old}(i,p)\boldsymbol{C}^*_{r,r_2,k,old}(j,p) \\
\qquad\quad + \sum_{r_1=1}^{r} \boldsymbol{C}_{r_1,r,k,old}(p,j)\boldsymbol{C}^*_{r_1,r,k,old}(p,i)
\end{cases} \tag{16}
$$

By defining $\zeta_{(i,j),r} \triangleq \sum_{k=1}^{K}\sum_{p=1,p\neq i}^{N}[a_{j,p,r}|\alpha_{r,(i,j)}|^2 + b_{i,j,p,r}\alpha^*_{r,(i,j)} + b^*_{i,j,p,r}\alpha_{r,(i,j)} + a_{i,p,r}]$, we actually express $\zeta_{(i,j),new}$ as summation of $R$ terms: $\zeta_{(i,j),new} = \sum_{r=1}^{R}\zeta_{(i,j),r}$, and each term $\zeta_{(i,j),r}$ is associated with only one parameter $\alpha_{r,(i,j)}$, $r = 1,\dots,R$. Therefore, we need only set the derivative of $\zeta_{(i,j),r}$ with respect to $\alpha^*_{r,(i,j)}$ to zero, for the estimation of $\alpha_{r,(i,j)}$. The solution is given below:

$$
\tilde{\alpha}_{r,(i,j)} = -\left(\sum_{k=1}^{K}\sum_{p=1,p\neq i}^{N} a_{j,p,r}\right)^{-1}\left(\sum_{k=1}^{K}\sum_{p=1,p\neq i}^{N} b_{i,j,p,r}\right) \tag{17}
$$

The L-stage constitutes sweeps over all the index pairs $(i, j)$, $1 \le j < i \le N$, as is given in (10), with the optimal parameters found with (17). When all the possible index pairs are exhausted once (one sweep), we switch to the U-stage that is accomplished similarly to the L-stage, with the only exception that the index pair satisfy $1 \le i < j \le N$. Then after one sweep for the U-stage we return to the L-stage. These 2 stages are alternated one after another until convergence.

### B. *Remarks and discussion*

In this subsection we provide discussions on the properties of the proposed GNJD algorithm, such as computation complexity and convergence, as well as some implementation remarks.

*Remark 1(Computation complexity per sweep):* From (10) and (17) we note that in each iteration for fixed index pair $(i, j)$ in L-stage, the load for calculating $\tilde{\alpha}_{r,(i,j)}$, $r = 1,\dots,R$ is $\mathcal{O}(2NKR^2)$, and those for updating $\boldsymbol{C}_{r_1,r_2,k}$ and $\boldsymbol{L}_{r,k}$ are $\mathcal{O}(NKR^2)$ and $NR$, respectively. Therefore, taking into consideration that there exist $N(N-1)/2$ such iterations in each sweep and that similar analysis holds for U-stage as well, the overall computation load per sweep of GNJD is $\mathcal{O}(3KR^2N^3)$. In addition, with similar analysis we note that the computation complexities for JNJD, NOJoB, and GOJD are $\mathcal{O}(6KRN^3)$, $\mathcal{O}(2KR^2N^3)$ and $\mathcal{O}(3KR^2N^3)$, respectively. We note here that the computation load per sweep of GNJD is comparable to that of GOJD, while JNJD is the least complex as it uses much less target matrices in the computation procedure.

*Remark 2(Convergence):* It is important to note that the transformation in (10) always reduces the cost function defined in (6) for each iteration. More exactly, if we take the L-stage for example, the cost function in the iteration with index pair $(i, j)$ is reduced to (15) since only the *ith* columns and rows of the target matrices are updated. Therefore, substituting the solution $\tilde{\alpha}_{r,(i,j)}$ into (15), we have the following result:

$$
\Delta\zeta \triangleq \zeta_{(i,j),new} - \zeta_{(i,j),old} = -\frac{\sum_{r=1}^{R}\sum_{k=1}^{K}\sum_{p=1,p\neq i}^{N}|b_{i,j,p,r}|^2}{\sum_{k=1}^{K}\sum_{p=1,p\neq i}^{N} a_{j,p,r}} \tag{18}
$$

where $a_{j,p,r}$, $b_{i,j,p,r}$ are defined in (16). We note that $\Delta\zeta$ is always non-positive. Since the cost function is lower-bounded by zero by definition, and is always reduced by the proposed iterations as is implied by (18), the proposed algorithm is guaranteed to at least converge to a local minimum.

*Remark 3(Parallelization for calculation of $\alpha_{r,(i,j)}$):* We note that (17) actually infers parallel calculation of $\alpha_{r,(i,j)}$ for all $r = 1,\dots,R$, instead of the dataset-wise sequential scheme that calculates $\alpha_{1,(i,j)}, \alpha_{2,(i,j)}, \dots, \alpha_{R,(i,j)}$ one after another. Noting that the above sequential updating of unmixing matrix is required for NOJoB, the proposed GNJD algorithm is expected to perform faster than NOJoB when handling large number of datasets.

In addition, parallelization over matrix dimensionality could be considered for further acceleration [30, 31]. More exactly, we can use the column-wise parallelization scheme [30] that calculates all $\alpha_{r,(i,j)}$'s with identical $i$ index yet distinct $j$ indices simultaneously. It is interesting to note that this scheme is able to largely reduce the number of rotations needed in one particular sweep without losing any accuracy of calculation in the NJD context, and thus similar property could be expected for GNJD. Other parallelization schemes, such as the tournament player ordering scheme [31], could also be used in our GNJD method.

*Remark 4(On the balance of intra-set statistics and inter-set statistics):* We note that the cost function for GNJD could be rewritten as the sum of two terms as follows:

$$\eta = \overbrace{\sum_{k=1}^{K}\sum_{r=1}^{R}\left\|\text{off}(\boldsymbol{B}_r \cdot \boldsymbol{C}_{r,r,k} \cdot \boldsymbol{B}_r^H)\right\|_F^2}^{\eta_1} + \overbrace{\sum_{k=1}^{K}\sum_{r_2=r_1+1}^{R}\sum_{r_1=1}^{R-1}\left\|\text{off}(\boldsymbol{B}_{r_1} \cdot \boldsymbol{C}_{r_1,r_2,k} \cdot \boldsymbol{B}_{r_2}^H)\right\|_F^2}^{\eta_2} \tag{19}$$

In J-BSS applications, $\eta_1$ involves off-norms of target matrices $\boldsymbol{C}_{r,r,k}$'s that are calculated with intra-set statistics, such as auto-covariance within the $r$th dataset, yet $\eta_2$ is related to matrices $\boldsymbol{C}_{r_1,r_2,k}$, $r_1 < r_2$ that are usually obtained from inter-set statistics across datasets, such as cross-covariance or cross 4th-order cumulant matrices. Details on the calculation of these target matrices for J-BSS could be found in subsection II.B. The goal of J-BSS is to: (1) estimate $R$ unmixing matrices $\boldsymbol{B}_r$'s simultaneously, and (2) at the same time align the permutations of columns for all $\boldsymbol{B}_r$'s, $r$ = 1,2,…,$R$ by minimizing $\eta$. It is important to note, with regards to the 2 terms $\eta_1$ and $\eta_2$, that minimizing $\eta_1$ only yields estimates of unmixing matrices without permutation alignment, yet minimization of $\eta_2$ achieves the same goal as J-BSS: estimation of unmixing matrices as well as permutation alignment. As such, the inter-set statistical target matrices actually play a more important role than the intra-set ones in achieving J-BSS, and this infers that we should give $\eta_2$ a larger portion in $\eta$ when performing GNJD.

In addition, the numbers of intra-set statistical and inter-set statistical target matrices in GNJD model (4) are $RK$ and $R(R-1)K/2$, respectively, and thus the portion of $\eta_2$ in $\eta$ is roughly $(R-1)/(R+1)$, which is small when $R$ takes small values. This indicates that using the cost function (19) might encounter some problem for small numbers of datasets with regards to permutation alignment. Indeed, a smaller portion of $\eta_2$ will result in a number of local minima in the manifold of the cost function that come from the intra-set statistics, and GNJD is likely to be stuck into a local minimum in such cases.

Indeed, we have observed with simulations quite a few non-optimal convergence patterns for NOJoB and GNJD when $R$ takes small values, which in return verifies our analysis above. It is important to note that GOJD does not have such problem as it uses inter-set statistics only.

To solve the above problem, we consider to remove intra-set statistical target matrices from GNJD when $R$ is smaller than $R'$ (empirically set to 5), and remain both intra-set statistics and inter-set statistics otherwise.

*Remark 5(Termination criteria):* Several stopping criteria are available such as monitoring the changes of the off-norm based cost function defined in (6) between 2 adjacent sweeps. Here we choose to terminate the iterations when the overall updates of $\boldsymbol{L}_r$ and $\boldsymbol{U}_r$ from all the elementary matrices $\boldsymbol{T}_{r,(i,j)}$ are sufficiently close to the identity matrix. That is to say, we shall stop the iterations if the following inequality is met:

$$\max_r \|\prod_{i,j}\boldsymbol{T}_{r,(i,j)}\prod_{i,j}\boldsymbol{T}'_{r,(i,j)} - \boldsymbol{I}_N\|_F \le \tau \tag{20}$$

where $\boldsymbol{T}_{r,(i,j)}$ and $\boldsymbol{T}'_{r,(i,j)}$ denote the elementary rotation matrices in the L and U stages, respectively, and $\tau$ is a preset threshold (e.g. we use $\tau = 10^{-6}$ in experiments).

*Remark 6 (Normalization):* It is important to note that the convergence of the proposed algorithm does not necessarily imply joint diagonalization for all the asymmetric NJD datasets, as the criterion (6) is not scale invariant, and this may result in inaccurate estimates of $\alpha_{r,(i,j)}$ especially when $N$ and $K$ are large [11]. In practice, an efficient way to alleviate this is normalization [10]. It is simply achieved by multiplying $\tilde{\boldsymbol{B}}_r$ from the right by a diagonal matrix $\boldsymbol{D}_r$ such that each row of $\tilde{\boldsymbol{B}}_r$ is of unit norm, and multiplying $\boldsymbol{C}_{r_1,r_2,k}$ by $\boldsymbol{D}_{r_1}^{-1}$ and $\boldsymbol{D}_{r_2}^{-H}$ on both sides. In addition, we note that the computation cost of normalization for each sweep is $\mathcal{O}(KR^2N^2)$ which is negligible compared to the overall complexity. Moreover, we have observed in our experiments that the proposed algorithm still converges without normalization. However, there is no theoretical proof for such observations, and we still use it as a proper precaution.

Based on the description of GNJD algorithm and remarks in Subsection III.A and III.B, we summarize the proposed GNJD algorithm in TABLE 1.

TABLE 1
IMPLEMENTATION OF PROPOSED ALGORITHM

- **Input:** $R(R+1)/2$ sets of target matrices $\boldsymbol{C}_{r_1,r_2,k}$, $r_1 = 1,...,R$, $r_2 = r_1,...,R$, $k = 1,...,K$, threshold $\tau$, (e.g. $\tau = 10^{-6}$), and an integer $R'$ (e.g. $R' = 5$)
- **Output:** $R$ estimated unmixing matrices $\boldsymbol{B}_r$ with aligned permutations
- **Implementation:**

  $\boldsymbol{B}_r \leftarrow \boldsymbol{I}_N$, $\gamma_{old} \leftarrow 0$, $\zeta \leftarrow \tau + 1$.
  **if** $R < R'$ **do**
      Exclude intra-set statistical target matrices $\boldsymbol{C}_{r,r,k}$ (*Remark 4*)
  **end**
  **while** $\zeta \ge \tau$ **do**
      The U-stage: $\boldsymbol{U}_r \leftarrow \boldsymbol{I}_N$
      **for all** $1 < j \le N$ **do**
          - Obtain optimal elementary upper-triangular matrices $\boldsymbol{T}'_{r,(i,j)}$ for $1 \le r \le R$, $1 \le i < j$ by (17), in parallel manner (*Remark 3*);
          - Update target matrices according to (10), and update matrices: $\boldsymbol{U}_r \leftarrow \boldsymbol{T}'_{r,(i,j)}\boldsymbol{U}_r$
      **end for**
      The L-stage: $\boldsymbol{L}_r \leftarrow \boldsymbol{I}_N$
      **for all** $1 \le j < N$ **do**
          - Obtain optimal elementary lower-triangular matrices $\boldsymbol{T}_{r,(i,j)}$ for $1 \le r \le R$, $j < i \le N$ by (17), in parallel manner (*Remark 3*);
          - Update target matrices according to (10) and update matrices: $\boldsymbol{L}_r \leftarrow \boldsymbol{T}_{r,(i,j)}\boldsymbol{L}_r$
      **end for**
      - $\boldsymbol{B}_r \leftarrow \boldsymbol{L}_r\boldsymbol{U}_r\boldsymbol{B}_r$, $\gamma_{new} \leftarrow \max_r \|\prod_{i,j}\boldsymbol{T}_{r,(i,j)}\prod_{i,j}\boldsymbol{T}'_{r,(i,j)} - \boldsymbol{I}_N\|_F$ (*Remark 5*)
      - $\zeta \leftarrow |\gamma_{new} - \gamma_{old}|$, $\gamma_{old} \leftarrow \gamma_{new}$
      - Normalize $\boldsymbol{B}_r$ and $\boldsymbol{C}_{r_1,r_2,k}$ according to *Remark 6*
  **end while**

## IV. EXPERIMENT RESULTS

In this section, we illustrate the performances of the proposed GNJD algorithm with 5 progressively more complex experiments (the list of compared algorithms may vary for different experiments and is specified at the beginning of each experiment). More exactly, we illustrate in Experiment 1 behaviors of compared algorithms with exactly diagonalizable matrices, with emphasis on the converging patterns. In Experiment 2, approximately jointly diagonalizable asymmetric target matrices are used in order to examine the performances in turbulences. In Experiment 3, we test the performance of compared algorithms in the context of 2nd-order J-BSS. In Experiment 4 and 5, results with real-world datasets are presented, wherein applications with electrocardiogram (ECG) mixtures and frequency domain speech mixtures are taken into consideration.

All the experiments are done under following configurations; CPU: Intel Core i7-4930MX 3.0GHz; Memory: 32GB; System: 64bit Windows 7; Matlab R2013b.

**Experiment 1. Performances with exactly diagonalizable matrices:** In this experiment, we demonstrate and compare the convergence behaviors of generalized JD algorithms including GNJD, NOJoB [26], JNJD [27] and GOJD [24]. We generate $K(R+1)R/2$ target matrices $\boldsymbol{C}_{r_1,r_2,k} \in \mathbb{C}^{N\times N}$, $1 \le r_1 \le r_2 \le R$, $1 \le k \le K$, according to (4) where both the real and imaginary parts of the elements of $\boldsymbol{A}_{r_1}, \boldsymbol{A}_{r_2} \in \mathbb{C}^{N\times N}$ and the diagonal elements of $\boldsymbol{D}_{r_1,r_2,k} \in \mathbb{C}^{N\times N}$ are drawn from normal distributions with zero mean and unit variance, $K$, $R$, and $N$ denote the number of target matrices in each (asymmetric) NJD set, number of mixing matrices to be estimated, and matrix dimensionalities, respectively. The overall off-norm (ORON) at the *ith* sweep is calculated to evaluate the converging procedure which is defined as follows:

$$ORON(i) = \frac{\sum_{r_1,r_2,k} \left\| \mathrm{off}(\boldsymbol{B}_{r_1}^{(i)} \cdot \boldsymbol{C}_{r_1,r_2,k} \cdot \boldsymbol{B}_{r_2}^{(i)H}) \right\|_F^2}{\sum_{r_1,r_2,k} \left\| \mathrm{diag}(\boldsymbol{B}_{r_1}^{(i)} \cdot \boldsymbol{C}_{r_1,r_2,k} \cdot \boldsymbol{B}_{r_2}^{(i)H}) \right\|_F^2} \quad (21)$$

where $\boldsymbol{B}_r^{(i)}$ denotes update of the *rth* unmixing matrix in the *ith* sweep, operations '$\mathrm{off}(\cdot)$' and '$\mathrm{diag}(\cdot)$' set the diagonal and off-diagonal elements of their entries to zero, respectively.

We perform GNJD and NOJoB upon the above generated target matrices. For JNJD, the target matrices are selected from those for GNJD by requiring $r_2 = r_1 + 1$. In addition, the target matrices for GOJD are $\boldsymbol{P}_{r_1}\boldsymbol{C}_{r_1,r_2,k}\boldsymbol{P}_{r_2}^H$ for all $1 \le r_1 \le r_2 \le R$, $1 \le k \le K$, where $\boldsymbol{P}_r$ is the pre-whitening matrix obtained from singular value decomposition (SVD) of $\boldsymbol{A}_r$. We draw the ORON curves of the compared algorithms versus the number of sweeps from 10 independent runs, under the following settings: (a) $K=20, N=5, R=3$; (b) $K=20, N=5, R=10$; (c) $K=20$, $N=5, R=15$; (d) $K=20, N=5, R=20$. The results are plotted in Fig. 1. We note here that in all the 4 settings, both GNJD and GOJD converge nicely within 15 sweeps, with superlinear converging pattern particularly when approaching the final solution. In contrary, we note that non-optimal converging patterns exist for NOJoB in settings (b), (c), and (d), which are at times observed to fail to converge into global minimum. This suggests that using NOJoB for large number of datasets is likely to have problems of permutation misalignment. In addition, JNJD almost completely fail to converge into global minimum in settings (b), (c), and (d). The above observations suggest that GOJD and the proposed GNJD algorithms are the most reliable ones among the competitors in achieving both estimation of loading matrices and permutation alignment, when handling small or large number of datasets. On the other hand, NOJoB provides quite reliable performance when the number of datasets is small, yet lacks some efficiency for large number of datasets as is indicated in the non-optimal converging patterns. JNJD is only able to handle small number of datasets.

It is important to note that in this experiment perfect pre-whitening is done for GOJD, noting that the matrices $\boldsymbol{P}_r$ are obtained with SVD of the true loading matrices $\boldsymbol{A}_r$, $r = 1, 2, \ldots, R$. However, in noisy cases where uncorrectable errors for subsequent GOJD stage are likely to be introduced in pre-whitening [23], the performance of GOJD will deteriorate, as will be shown in following experiments.

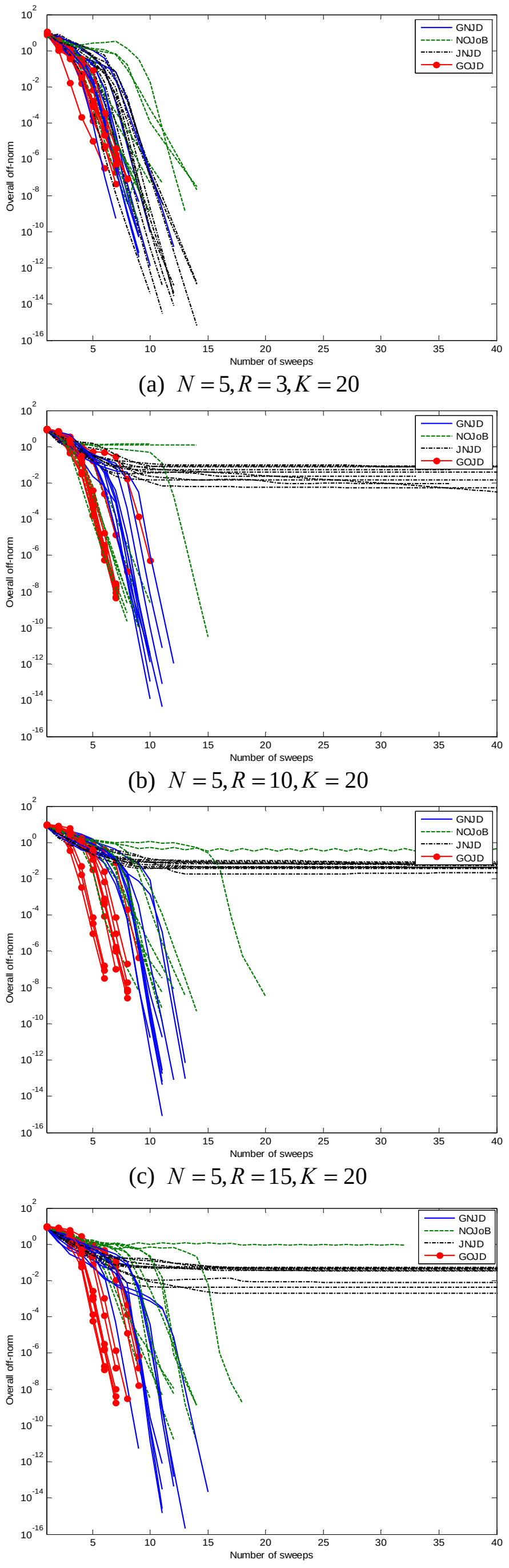

(a) $N=5, R=3, K=20$

(b) $N=5, R=10, K=20$

(c) $N=5, R=15, K=20$

(d) $N=5, R=20, K=20$

Fig. 1. Overall off-norm (ORON) of GNJD, NOJoB, JNJD, and GOJD versus the number of sweeps

**Experiment 2. Performances with approximately jointly diagonalizable matrices:** In this experiment, we examine and compare the performances of generalized JD algorithms (GNJD, NOJoB, GOJD, JNJD) with approximately jointly diagonalizable target matrices generated as follows:

$$\boldsymbol{C}'_{r_1,r_2,k} = \sigma_s \frac{\boldsymbol{C}_{r_1,r_2,k}}{\|\boldsymbol{C}_{r_1,r_2,k}\|_F} + \sigma_n \frac{\boldsymbol{N}_{r_1,r_2,k}}{\|\boldsymbol{N}_{r_1,r_2,k}\|_F} \tag{22}$$

where $\boldsymbol{C}_{r_1,r_2,k}$ is constructed in the same way as Experiment 1, $\boldsymbol{N}_{r_1,r_2,k}$ is the noise term with both the real and imaginary parts drawn from normal distributions with zero mean and unit variance, and $\sigma_s$, $\sigma_n$ denote the levels of signal and noise, respectively. In addition, we define signal-to-noise ratio (SNR) in this case as:

$$SNR = 10\log_{10}(\sigma_s / \sigma_n) \tag{23}$$

We perform GNJD and NOJoB upon $\boldsymbol{C}'_{r_1,r_2,k}$, $r_1 = 1,2,...,R$, $r_2 = r_1,...,R$, $k = 1,2,...,K$, and JNJD upon the subset of $\boldsymbol{C}'_{r_1,r_2,k}$ with $r_2 = r_1 + 1$. In addition, the true loading matrices $\boldsymbol{A}_r$ are assumed known for GOJD to facilitate pre-whitening. The target matrices for GOJD are $\boldsymbol{P}_{r_1}\boldsymbol{C}'_{r_1,r_2,k}\boldsymbol{P}^H_{r_2}$ with pre-whitening matrices $\boldsymbol{P}_r$ obtained from SVD of $\boldsymbol{A}_r$, $r = 1, 2, \ldots, R$.

We evaluate the performances of all compared algorithms by the joint inter-symbol-interference (J-ISI) [24] defined as:

$$\text{J-ISI}(\boldsymbol{G}) \triangleq \frac{1}{2N(N-1)}\left[\sum_{i=1}^{N}\left(\sum_{j=1}^{N}\frac{g_{ij}}{\max_k g_{ik}} - 1\right) + \sum_{j=1}^{N}\left(\sum_{i=1}^{N}\frac{g_{ij}}{\max_k g_{kj}} - 1\right)\right] \tag{24}$$

where $g_{ij}$ denotes the $(i, j)th$ element of $\boldsymbol{G} \triangleq \sum_{r=1}^{R}|\boldsymbol{W}_r\boldsymbol{A}_r|$, $|\cdot|$ calculates the absolute value of each element of its entry, $\boldsymbol{W}_r$ denotes the $rth$ normalized estimated unmixing matrix, and $\boldsymbol{A}_r$ is the $rth$ normalized true mixing matrix. We note here that J-ISI takes into account both the accuracy and the permutation of the rows of the estimates of unloading matrices, and a small value of it indicates an accurate estimate of each individual unmixing matrix, as well as nicely aligned permutations for all the unmixing matrices.

We fix $\sigma_n$ to 0.01, let $SNR$ vary from 0 – 20dB, and plot the J-ISI curves obtained from 100 independent runs in Fig. 2 under the following two settings: (a) $K = 20, N = 5, R = 5$; (b) $K = 20, N = 5, R = 20$. It is shown that GNJD provides the best results when SNR is lower than 16dB, followed by NOJoB and then by GOJD in both cases. This observation clearly indicates the pros and cons of non-orthogonal and orthogonal GJD that the former is likely to outperform the latter in the presence of additive turbulences, and this coincides nicely with the comparison of the non-orthogonal and the orthogonal in the JD context [32]. Moreover, it is interesting to note that JNJD provides competitive performance when $R = 5$, which again suggests its applicability for small number of datasets.

**Experiment 3. 2nd-order J-BSS of synthetic multi-set data:** In this experiment, we compare the proposed GNJD algorithm with NOJoB, GOJD, and MCCA in 2nd-order J-BSS of synthetic multi-set data (JNJD is excluded due to its poor performance in the following settings). The non-stationary sources are generated as:

$$\boldsymbol{s}_n(t) = \boldsymbol{\Pi}\boldsymbol{s}'_n(t) \tag{25}$$

where $\boldsymbol{s}_n(t) \triangleq [\, s_n^{(1)}(t), s_n^{(2)}(t), ..., s_n^{(R)}(t)]^T \in \mathbb{C}^R$, $s_n^{(r)}(t)$ denotes the $nth$ source at time instance $t$ in the $rth$ dataset, and $\boldsymbol{\Pi} \in \mathbb{C}^{R\times R}$ is a full rank matrix used to introduce inter-set correlations. Both the real and imaginary parts of each element of $\boldsymbol{\Pi}$ are drawn from normal distribution of zero mean and unit variance. $\boldsymbol{s}'_n(t) \triangleq [\, s_n'^{(1)}(t), ..., s_n'^{(R)}(t)]^T$ are complex BPSK that are amplitude modulated across time slots:

$$s_n'^{(r)}(t) = \sum_{m=1}^{M} \eta_m s_{n,m}'^{(r)}(t) \tag{26}$$

where $\eta_m$ 's are randomly drawn from uniform distribution over [0, 1] and $s_{n,m}'^{(r)}(t)$ is defined as:

$$s_{n,m}'^{(r)}(t) = \begin{cases} b^{(m)}(t - m(1-\alpha)L), & m(1-\alpha)L < t \le m(1-\alpha)L + L \\ 0 & , \; else \end{cases} \tag{27}$$

where $b^{(m)}(t)$ is a BPSK signal at time $t$ with value selected from symbols $[1+\vec{\mathrm{i}},\ 1-\vec{\mathrm{i}}]$ with equal probability, $L$ is the number of samples of $b^{(m)}(t)$, and $\alpha$ denotes the overlapping rate of parts in $b^{(m-1)}(t)$ and $b^{(m)}(t)$ that contribute to the same time duration in $s_n'^{(r)}(t)$. The mixtures are constructed as:

$$\boldsymbol{x}_r(t) = \sigma_s \frac{\boldsymbol{A}_r\boldsymbol{s}^{(r)}(t)}{\|\boldsymbol{A}_r\boldsymbol{s}^{(r)}(t)\|_F} + \sigma_n \frac{\boldsymbol{n}^{(r)}(t)}{\|\boldsymbol{n}^{(r)}(t)\|_F}, \quad r = 1,2,...,R \tag{28}$$

where both the real and imaginary parts of mixing matrices $\boldsymbol{A}_r$'s are taken randomly from normal distribution with zero mean and unit variance, and $\boldsymbol{n}^{(r)}(t)$ is the noise term in the $rth$ dataset. The spatial correlations for noise terms associated with distinct datasets are introduced similarly to (25). $\sigma_s$ and $\sigma_n$ denote the signal and noise levels respectively. We note here that $s_n'^{(r)}(t)$ (see Fig. 3 as an example) are short-time stationary, and thus target matrices $\boldsymbol{C}_{r_1,r_2,k}$ are constructed according to (5). In practice, we calculate the sampled version of $\boldsymbol{C}_{r_1,r_2,k}$ as:

$$\tilde{\boldsymbol{C}}_{r_1,r_2,k} = \frac{1}{L'}\sum_{t=1}^{L'} \boldsymbol{x}_{r_1,k}(t)\boldsymbol{x}_{r_2,k}^H(t) \tag{29}$$

where $\boldsymbol{x}_{r,k}(t)$ denotes the $kth$ block segmented from $\boldsymbol{x}_r(t)$ along the time dimension, with block length $L'$, and overlapping rate $\alpha' \in [0,1]$. SNR in this scenario is defined with $\sigma_s$ and $\sigma_n$ via (23).

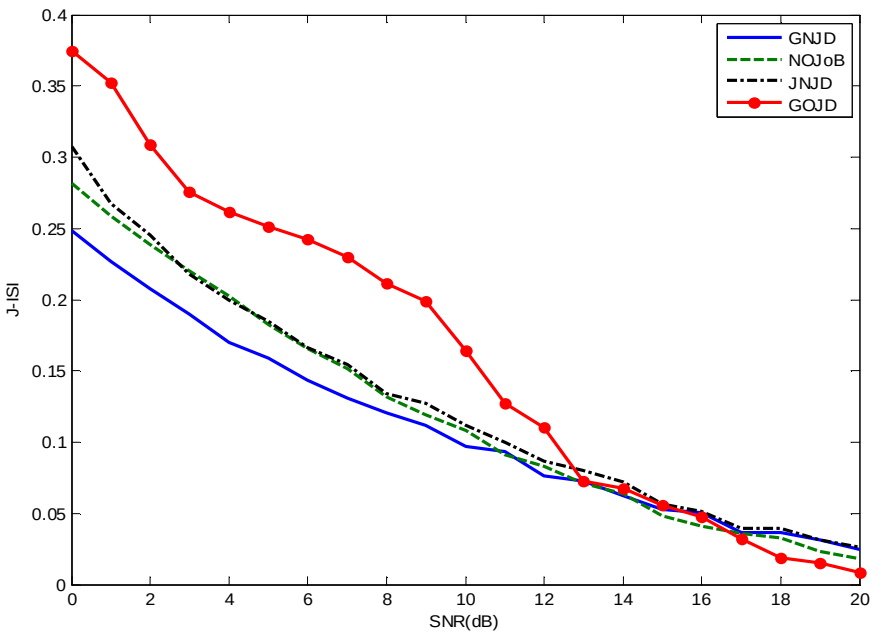

(a) $N = 5, R = 5, K = 20$

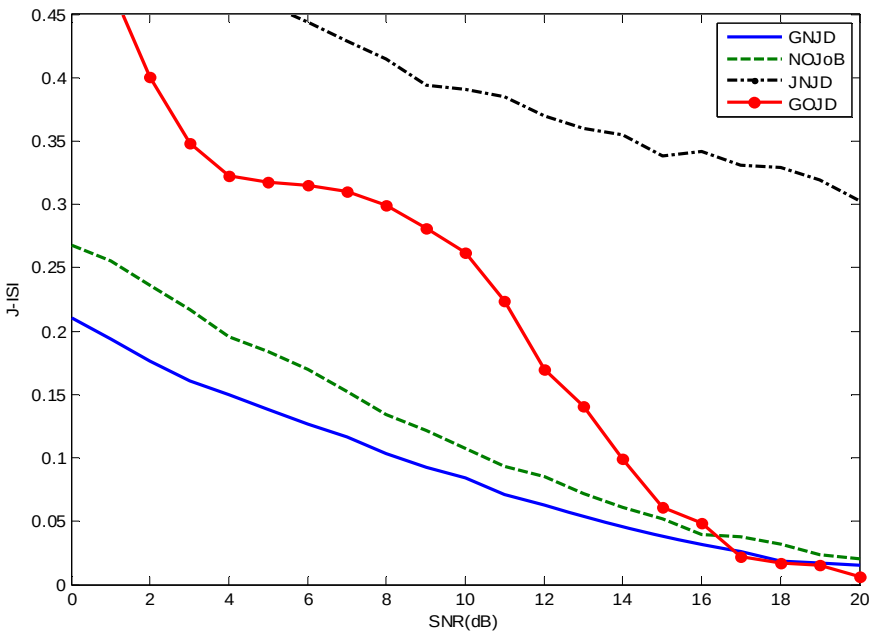

(b) $N = 5, R = 20, K = 20$

Fig. 2. J-ISI of GNJD, NOJoB, JNJD, GOJD in additive turbulences.

The target matrices for GOJD are obtained with pre-whitened datasets. The pre-whitening matrices are obtained via SVD of the sampled $\boldsymbol{x}_r(t)$. The target matrices for MCCA are cross-covariance matrices (without taking into account the temporal non-stationarity) across every pair of pre-whitened datasets.

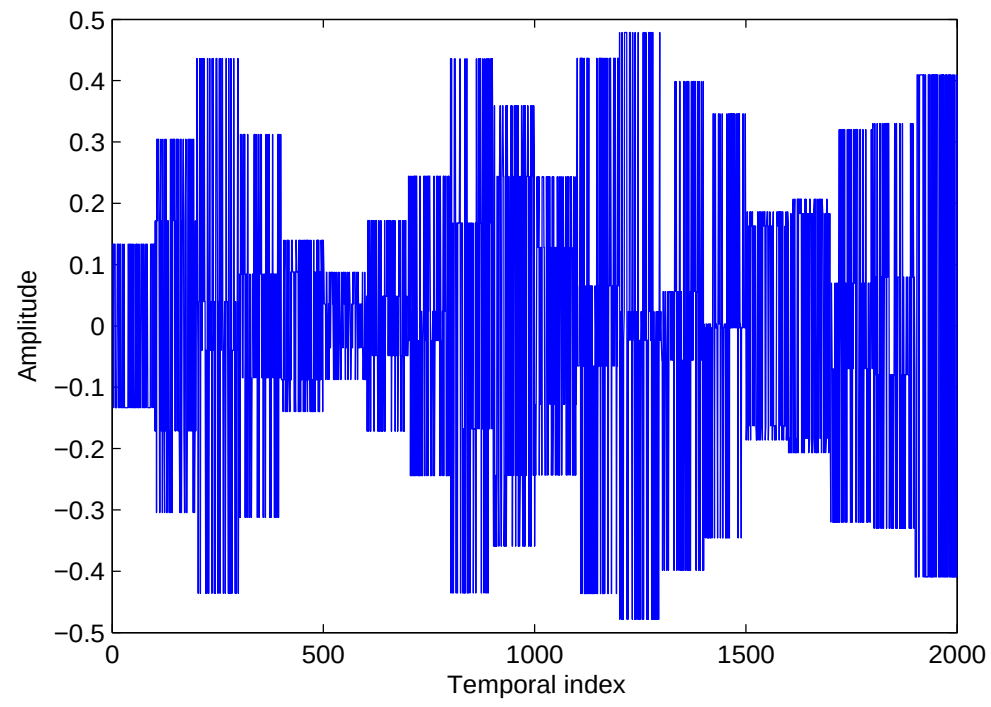

Fig. 3. The real part of an amplitude modulated BPSK signal with $L$ = 200, and $\alpha$ = 0.5

We fix the number of temporal samples $T = 2000$, $L = 200$, $L' = 100$, and $\alpha = \alpha' = 0.5$, and let SNR vary from -2dB – 10dB. The J-ISI curves versus SNR are plotted in Fig. 4 with contributions from 100 independent runs under the following 2 settings: (a) $N = 5, R = 10$; (b) $N = 5, R = 20$.

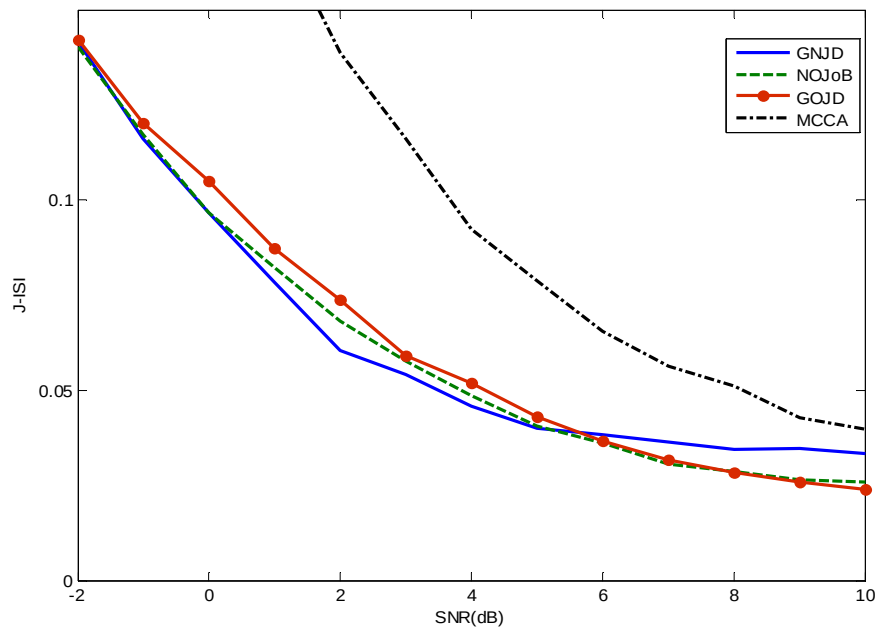

(a) $N = 5, R = 10$

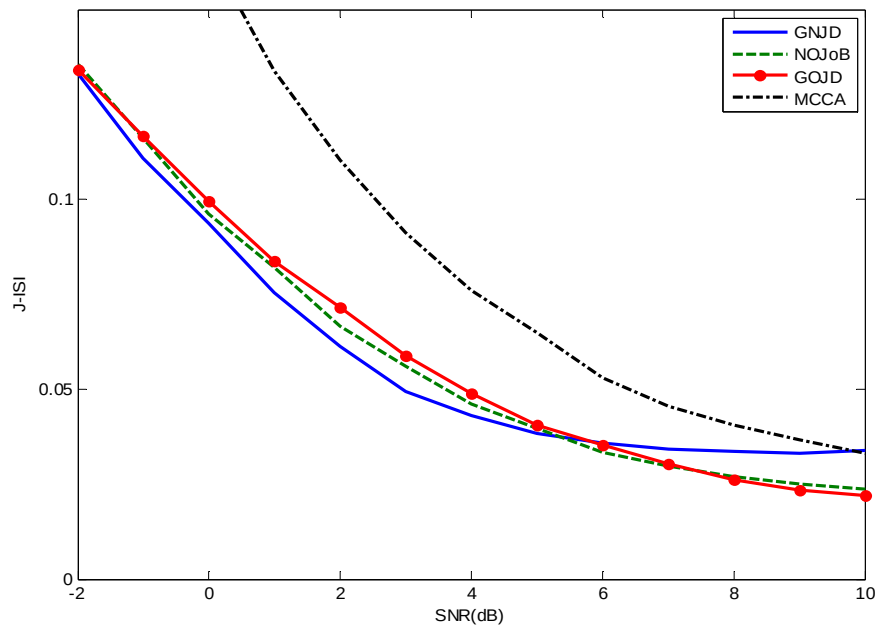

(b) $N = 5, R = 20$

Fig. 4. J-ISI of GNJD, NOJoB, GOJD, and MCCA in 2nd-order J-BSS

From Fig. 4 we see that GNJD yields best performance in low SNR levels (-2 – 5dB), followed by NOJoB, and GOJD in both scenarios. When SNR exceeds 5dB, the proposed GNJD algorithm slightly underperforms NOJoB and GOJD, but is still able to provide quite precise estimates (J-ISI is below 0.05). In addition, the merit of GNJD in low SNR is seen clearer when $R$ = 20, suggesting that the proposed GNJD algorithm is more advantageous for handling larger number of datasets.

We note in this experiment (as well as Experiment 2) that GNJD exhibits slightly lower performance than GOJD and NOJoB for high SNR's. That is because the latter two algorithms impose some constraints in the optimization (e.g. the orthogonality constraint for GOJD, the unit-norm constraint for rows of demixing matrix in NOJoB), and these constraints bring some merits over the unconstrained GNJD algorithm when SNR is high (such that prewhitening is precise for GOJD and non-optimal convergence is rarely encountered for NOJoB). In the presence of low SNR, however, that GOJD suffers from imprecise prewhitening and NOJoB encounters non-optimal convergence, GNJD is shown to yield best performance. In fact, noise is always present and can sometimes be quite high in real-world problems, and that makes GNJD particularly interesting in solving practical noisy problems.

**Experiment 4. Fetal electrocardiogram (ECG) separation:** In this experiment, we consider and compare 2nd-order J-BSS with GNJD, JNJD, GOJD, NOJoB, and MCCA in the context of fetal ECG separation with real-world 8-channel ECG data as is shown in Fig. 5, collected from a pregnant woman and made available in [33]. The sampling rate is 250Hz and 2500 samples (10s) are recorded.

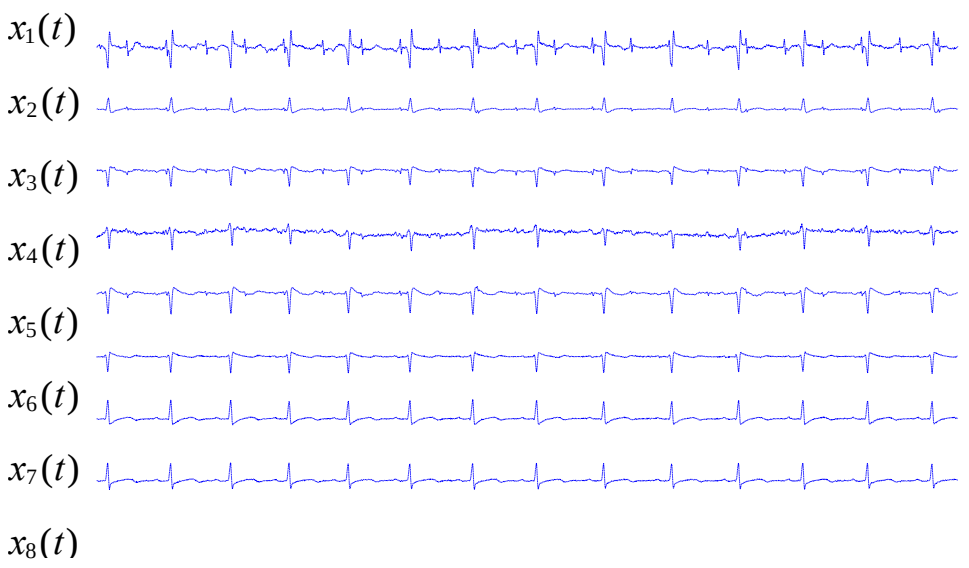

Fig. 5. The 8-channel ECG data from a pregnant woman

We note that the 2nd-order J-BSS with GOJD has already been applied to the same dataset and more details about the motivation of using J-BSS in such problems could be found in [24]. Here, we emphasize on the comparison of J-BSS algorithms in this application. The multi-set data is constructed following a similar procedure to [24] as:

$$\boldsymbol{x}_r(t) = [x_r(t), x_{r+1}(t), x_{r+2}(t), x_{r+3}(t)]^T, \quad r = 1,2,3,4,5 \tag{30}$$

The target matrices for GNJD and NOJoB are non-stationary covariance matrices, obtained similarly to (29) in Experiment 3 with block length $L' = 200$, and overlapping rate $\alpha' = 0.5$. The target matrices for JNJD and GOJD are constructed from those of GNJD via a similar procedure to that in Experiment 3. The target matrices for MCCA are cross-covariance matrices across every pair of pre-whitened datasets. The separation results are plotted in Fig. 6- Fig. 10 where those labelled $\boldsymbol{y}_r(t)$ denote the *rth* set of estimated sources, $r = 1,\ldots,5$.

From those figures we see that the strong and slow mother ECG could be nicely extracted by all the 5 compared algorithms, if we note the last two columns of the displayed results in Fig. 6- Fig. 10. In addition, the fetal ECG could be successfully restored in some components of the estimated results as well, if we look into the 1st components in $\boldsymbol{y}_1(t)$, $\boldsymbol{y}_2(t)$, and $\boldsymbol{y}_3(t)$ for all the compared algorithms. However, differences could be observed in other estimated results. In particular, we note that GNJD extracts more components related to fetal ECG than others if we compare the 2nd column of the displayed results, where GOJD, JNJD, NOJoB and MCCA only generate interferences while GNJD extracts fetal ECG components in $\boldsymbol{y}_1(t)$ and $\boldsymbol{y}_2(t)$. Generally, fetal ECG's are with particular interests in this application.

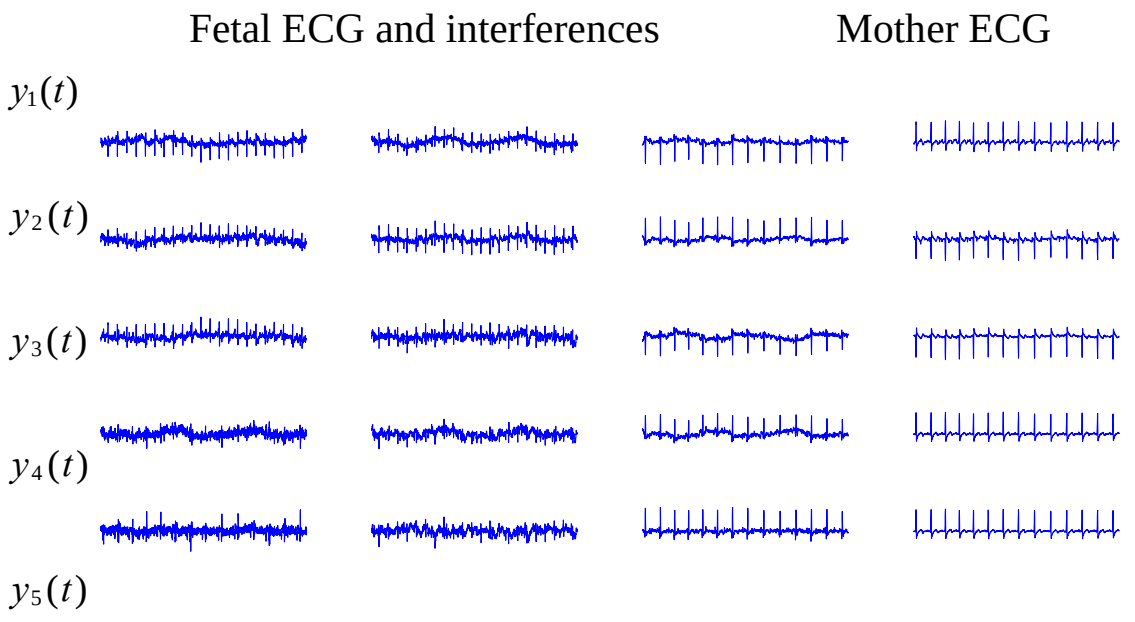

Fig. 6. Results from J-BSS with GNJD

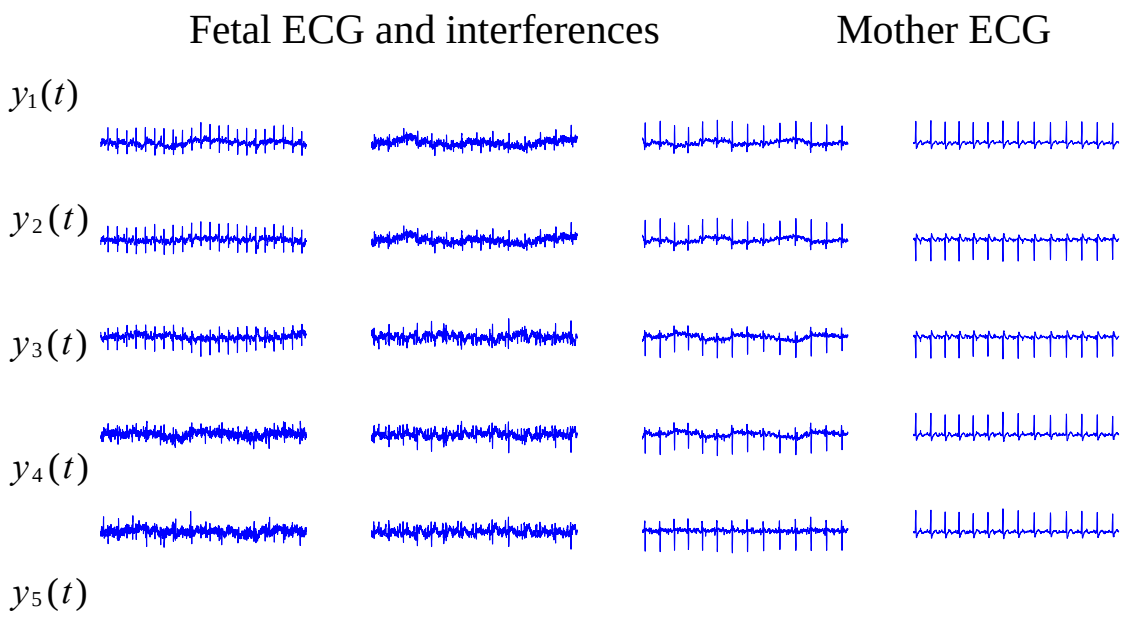

Fig. 7. Results from J-BSS with GOJD

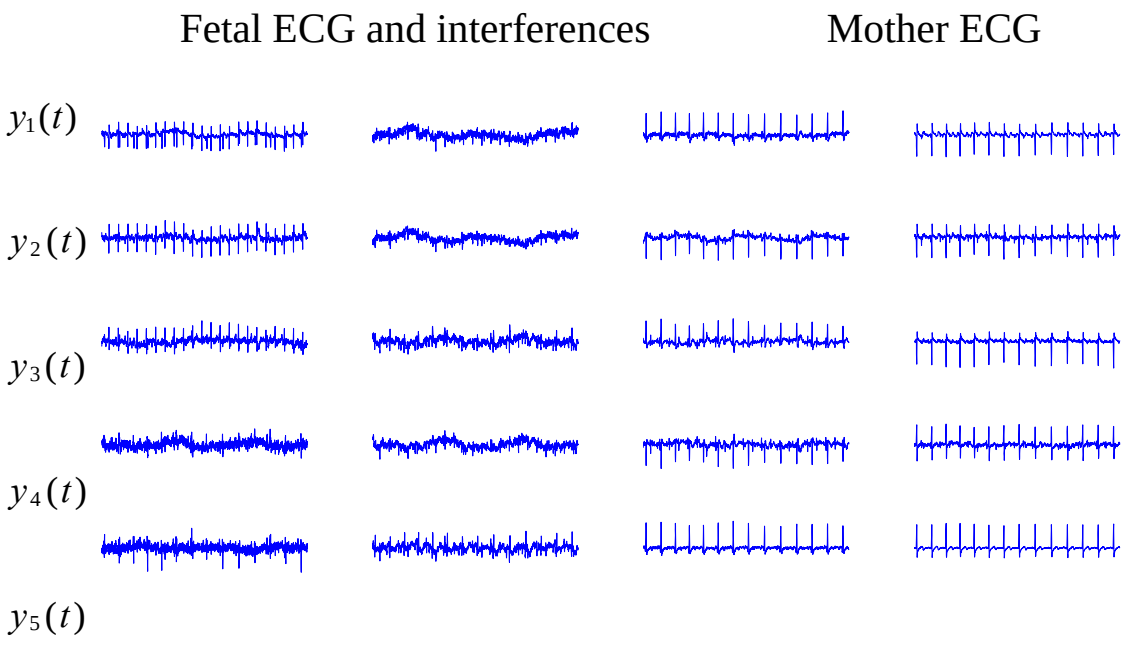

Fig. 8. Results from J-BSS with JNJD

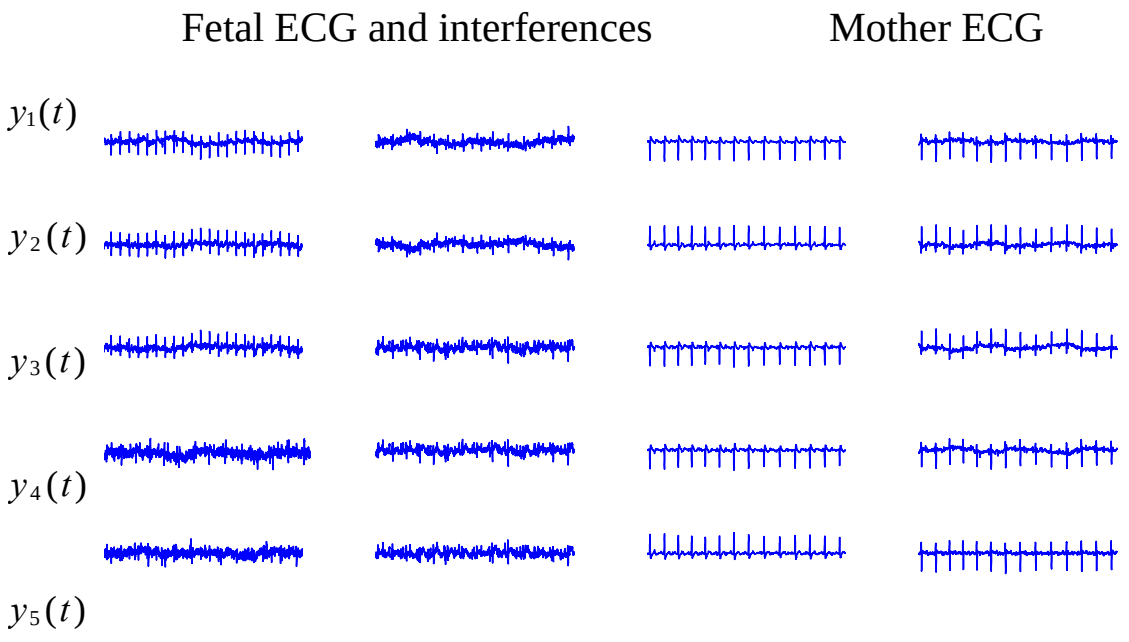

Fig. 9. Results from J-BSS with NOJoB

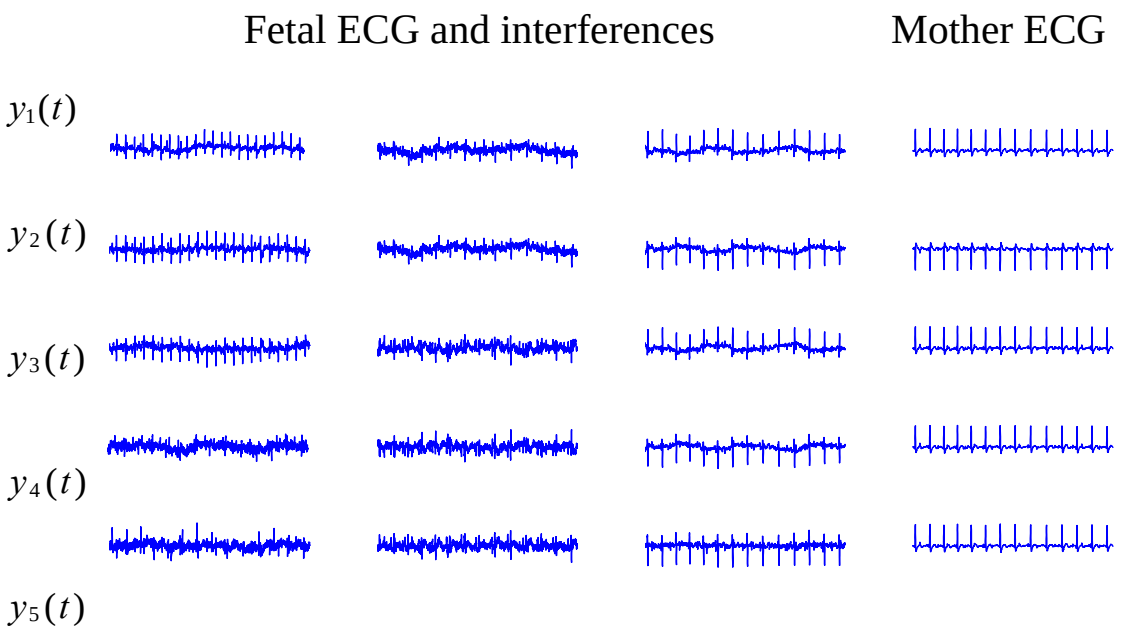

Fig. 10. Results from J-BSS with MCCA

**Experiment 5. Speech separation in frequency domain:** In this experiment, we consider and compare the applications of 2nd-order J-BSS with GNJD, JNJD, GOJD, NOJoB and MCCA to real-world speech separation in the frequency domain. We also include in the comparison some JD algorithms such as Cardoso's Jacobi-like OJD algorithm [1], Tichavsky and Yeredor's uniformly weighted exhaustive diagonalization by Gaussian iteration (UWEDGE) [6], complex-valued joint diagonalization via givens and shear rotations (C-JDi) [13] proposed by Mesloub, Abed-Meraim, and Belouchrani, and LU decomposition based complex-valued JD (LUCJD) [12]. We consider the scenario that two microphones receive two speeches. The real-world mixtures are obtained from SISEC2010 website [34] (scenarios Room 4 and Room 5). The room is a chamber with cushion walls of size $4.45m \times 3.55m \times 2.5m$, microphones are placed around the center of the room with height 1.25*m* with interspacing of 5.7*cm*, the sources are placed at height 1.25*m*, with distance of 1*m* from microphones in Room 4 setting, and 1.8m from microphones in Room 5 setting. The detailed description of the scenarios could be found in [34], and an illustration of Room 4 is given in Fig. 11.

Frequency domain speech mixtures are obtained with short time Fourier transform (STFT). The STFT frames are of length $F$ ($F$ = 2048 or 4096 in particular) for the competitors, half-overlapped with neighboring ones and windowed with sine function. The signals for the first $N_{bin} = F/2+1$ frequency bins are selected for the compared algorithms, as the rest of the frequency bins are redundant due to the symmetry of Fourier transform. The Matlab code for the above procedure could be found in E. Vincent's website [35].

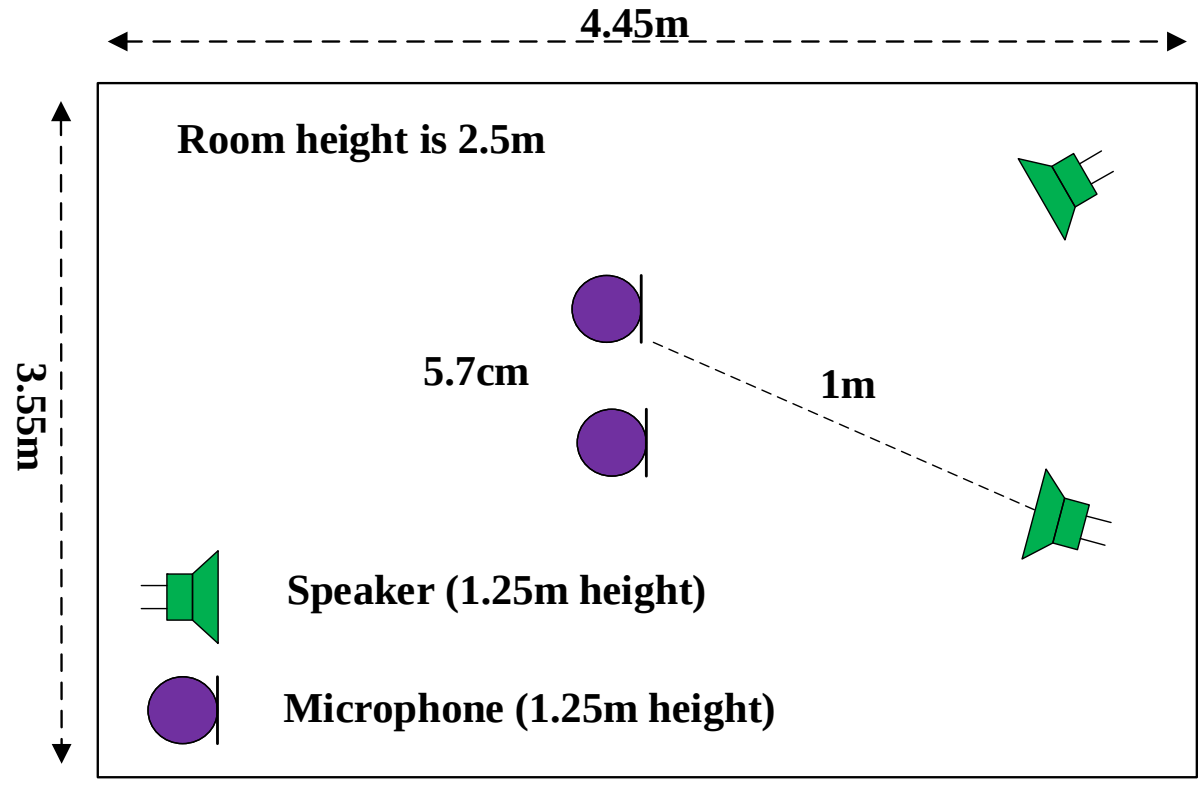

Fig. 11. Illustration of the simulated room settings

In addition, we construct $\lfloor N_{bin} / R \rfloor$ groups each containing $R$ adjacent frequency bins, and perform 2nd-order J-BSS described in Subsection II.B with GNJD, GOJD, JNJD, NOJoB, and MCCA for each group. For JD algorithms including OJD, UWEDGE, C-JDi, and LUCJD, 2nd-order BSS is performed for each frequency bin wherein the target matrices are obtained as non-stationary covariance matrices [3]. Moreover, pre-whitening is done at each frequency bin for OJD, MCCA, and GOJD.

After the above 2nd-order J-BSS (BSS) stage for each frequency bin, we calculate cross-covariance's of amplitudes of separated signal components of adjacent frequency bins, and admit those with larger cross-covariance's as coming from the same source. This amplitude-covariance based permutation alignment scheme is then performed sequentially to cover all the frequency bins to tackle the permutation ambiguity problem. In addition, the scaling ambiguity problem is solved via minimal distortion principle [36]. It is important to note that there actually exist more advanced permutation alignment schemes in the open literature [37]. However, we emphasize herein the separation performance for each frequency bin, and thus only use the basic one. Finally, inverse STFT is done to transform the separation results back to time domain.

We obtain the source speeches from Sawada's website [38]. The performance is evaluated with signal-to-interference ratio (SIR) and signal-to-distortion ratio (SDR). Detailed definitions and Matlab codes for these metrics could be found in [35].

All the compared algorithms are performed with STFT length $F$ = 2048 and $F$ = 4096. In addition, J-BSS algorithms (GNJD, NOJoB, GOJD, JNJD, MCCA) are performed with $R$ = 3 and $R$ = 4. Over all the above parameter options, the best result (with largest average SIR value) for each individual algorithm is selected and illustrated in TABLE **2** and TABLE **3** for Room 4 and 5, respectively.

We highlight the largest value along each column with bold font. From these tables we firstly observe that GNJD generates the best average results over the 2 sources in both scenarios, with regards to both SIR and SDR values. Moreover, for SIR values of each specific source, GNJD provides the best results for both sources in setting Room 4, and best SIR value for source 2 in setting Room 5. In addition, we note that GNJD is able to yield nice SIR's for both sources, in contrary to some other algorithms that only yield nice estimate for one source yet poor one for the other (e.g. NOJoB and GOJD in setting Room 5). Noting that SIR generally evaluates the separatability of the compared algorithms, the superiority of GNJD over all other competitors with regards to SIR clearly indicates its advantages in this aspect. Furthermore, we note that GNJD is able to provide quite competent SDR values (always ranked among top 3 for each specific source in both settings, and ranked 1st for the average value), indicating that GNJD could provide nice quality of the resulting speech estimates in addition to nice separatability.

TABLE 2
SIR AND SDR VALUES OF ALL COMPARED ALGORITHMS FOR SEPARATING REAL-WORLD SPEECH MIXTURES UNDER SCENARIO ROOM 4

| Method | Results for source 1 | | Results for source 2 | | Average results | |
|---|---|---|---|---|---|---|
| | $SIR_1$ | $SDR_1$ | $SIR_2$ | $SDR_2$ | $SIR_{ave}$ | $SDR_{ave}$ |
| GNJD | **18.99** | 12.25 | **19.39** | 11.76 | **19.19** | **12.00** |
| NOJoB | 8.60 | 5.13 | 15.07 | 10.07 | 11.83 | 7.60 |
| GOJD | 8.92 | 5.67 | 16.44 | 10.24 | 12.69 | 7.96 |
| JNJD | 12.94 | 8.18 | 16.97 | 10.48 | 14.95 | 9.33 |
| MCCA | 11.31 | 9.21 | 6.56 | 5.81 | 8.94 | 7.51 |
| OJD | 12.24 | 7.77 | 18.13 | 11.41 | 15.18 | 9.59 |
| UWEDGE | 18.77 | **13.11** | 11.60 | 7.76 | 15.18 | 10.43 |
| C-JDi | 10.29 | 7.36 | 18.29 | **11.99** | 14.29 | 9.68 |
| LUCJD | 18.18 | 12.78 | 11.52 | 7.80 | 14.85 | 10.29 |

TABLE 3
SIR AND SDR VALUES OF ALL COMPARED ALGORITHMS FOR SEPARATING REAL-WORLD SPEECH MIXTURES UNDER SCENARIO ROOM 5

| Method | Results for source 1 | | Results for source 2 | | Average results | |
|---|---|---|---|---|---|---|
| | $SIR_1$ | $SDR_1$ | $SIR_2$ | $SDR_2$ | $SIR_{ave}$ | $SDR_{ave}$ |
| GNJD | 14.40 | 5.78 | **21.27** | **11.61** | **17.84** | **8.70** |
| NOJoB | 4.67 | 1.47 | 18.96 | 10.70 | 11.82 | 6.08 |
| GOJD | 14.86 | 8.18 | 5.29 | 1.95 | 10.11 | 5.27 |
| JNJD | 9.91 | 5.21 | 17.33 | 9.83 | 13.62 | 7.52 |
| MCCA | 7.15 | 3.97 | 1.67 | -0.59 | 4.41 | 1.69 |
| OJD | 9.36 | 5.03 | 18.41 | 10.30 | 13.89 | 7.67 |
| UWEDGE | **17.20** | **10.25** | 9.29 | 3.92 | 13.24 | 7.09 |
| C-JDi | 8.46 | 4.09 | 17.27 | 9.92 | 12.87 | 7.00 |
| LUCJD | 9.40 | 4.55 | 17.48 | 9.74 | 13.44 | 7.14 |

The results clearly demonstrate the superiority of GNJD over other (generalized) JD variants in frequency domain based speech separation, thanks to its strong and robust performance for exploiting inter-set covariances of multi-set data in highly noisy environments.

## V. CONCLUSION

In this study, we considered the generalized non-orthogonal joint diagonalization (GNJD) of multiple asymmetric NJD datasets of which every pair share one common loading matrix. We proposed an algorithm for such problem based on LU decompositions and successive rotations. We have shown that the GNJD formulization could be obtained from the multi-set data model by using 2nd-order statistics, and thus the proposed algorithm could be used in multi-set data analysis applications such as joint blind source separation (J-BSS) and multi-modal /multi-subject data fusion. In addition, we have provided some theoretical analysis including complexity and convergence, as well as some implementation remarks.

Experiments are conducted to compare the proposed GNJD algorithm with existing ones of similar type, namely non-orthogonal joint blind source separation (NOJoB), generalized orthogonal joint diagonalization (GOJD), joint NJD (JNJD), and multiple canonical correlation analysis (MCCA), with artificial target matrices in both exactly and approximately jointly diagonalizable cases (Experiments 1 and 2), 2nd-order J-BSS applications over synthetic datasets (Experiment 3), and real-world J-BSS applications such as fetal ECG extraction (Experiment 4), and frequency domain speech separation (Experiment 5). The results generally show that GNJD is able to provide better performance when compared with NOJoB, GOJD, JNJD, and MCCA.

The sensitivities of GNJD and other generalized JD methods to difficult conditions are not studied in this paper and will be one of our future focuses.

## Acknowledgements

The authors would like to express sincere gratitude to Dr. Marco Congedo for providing us with source programs of NOJoB algorithm in [26]. We would also like to thank Prof. Karim Abed-Meraim, and his student Ammar Mesloub for providing us with source programs of C-JDi algorithm in [13].

The authors are grateful for the anonymous reviewers for their helpful comments and suggestions.

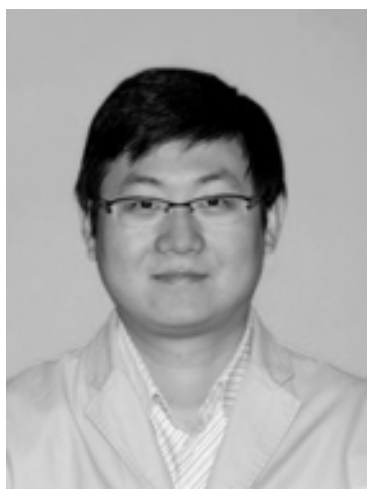

**Xiao-Feng Gong** (M'10) was born in the city of Yichang, Hubei province, China, in 1981. In 2003, he earned the B.S. degree (Information engineering) from Beijing Institute of Technology, China. From 2003 to 2009, he took the combined master and Ph.D programme of Beijing Institute of Technology, and earned the Ph.D degree (Communication and information system) in 2009. His Ph. D thesis concerned polarization sensitive array processing with tensors and hypercomplex numbers.

He was a lecturer from 2009 to 2012, and has been an associate professor since 2013, both with Dalian University of Technology, Dalian, China. He is currently a visiting research associate with KU Leuven Kortrijk campus, Kortrijk, Belgium. His research interests include multilinear algebra and its applications in blind source separation and array processing.

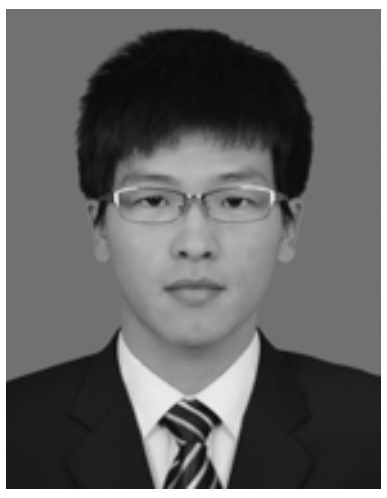

**Xiu-Lin Wang** was born in Dezhou city, Shandong province, China, in 1990. In 2012, he earned the B.S. degree (Electronic and information engineering) from Shan dong University Weihai campus, China. He is currently pursuing his M.S. degree in Dalian University of Technology, Dalian, China.

His research interests are algebraic methods for multiset signal processing.

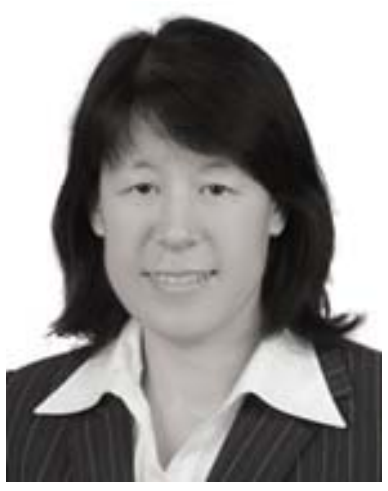

**Qiu-Hua Lin** (M'10) received the B.S. degree in wireless communication, the M.S. degree in communications and electronic systems, and the Ph.D. in signal and information processing, all from Dalian University of Technology, Dalian, China.

She is currently a Professor at the School of Information and Communication Engineering, Dalian University of Technology, Dalian, China. Her research interests include stochastic signal processing, array signal processing, biomedical signal processing, and image processing.